# Enhancing the Performance of Convolutional Neural Networks on Quality Degraded Datasets

Jonghwa Yim
Department of Computer Engineering
Ajou University
Suwon, Republic of Korea
jonhwayim@ajou.ac.kr

Kyung-Ah Sohn
Department of Software and Computer Engineering
Ajou University
Suwon, Republic of Korea
kasohn@ajou.ac.kr

*Abstract*—Despite the appeal of deep neural networks that largely replace the traditional handmade filters, they still suffer from isolated cases that cannot be properly handled only by the training of convolutional filters. Abnormal factors, including real-world noise, blur, or other quality degradations, ruin the output of a neural network. These unexpected problems can produce critical complications, and it is surprising that there has only been minimal research into the effects of noise in the deep neural network model. Therefore, we present an exhaustive investigation into the effect of noise in image classification and suggest a generalized architecture of a dual-channel model to treat quality degraded input images. We compare the proposed dual-channel model with a simple single model and show it improves the overall performance of neural networks on various types of quality degraded input datasets.

## I. Introduction

Recently, emerging convolutional neural networks (CNNs) have outpaced previous approaches in many problems in computer vision, including image classification, object detection, and object segmentation problems. The deep neural network model became possible due to powerful GPUs, which allow us to stack deep layers and process various features from the image data. Owing to the high performance of convolutional networks, the deep neural network model is applied to many practical problems; some of these require high confidence and extremely low error rates. This includes the real-time classification of autonomous cars or face recognition in a security system. Until now, the existing imaging solutions cannot be used to replace human resources completely due to unexpected errors. Despite the current achievements in the use of deep convolutional networks for image classification or object detection tasks, and that theoretically these networks outpace human recognition accuracy, their use in real-world applications is error-prone as their real performance is yet to be confirmed and particularly on quality degraded input images. Moreover, in previous research [1], the performance of the existing popular convolutional network model was severely affected by quality degraded input images, even if it was not visible. There are many factors that may have a negative impact on the existing network model—noise in the image, JPEG compression loss, blurred image, etc. These problems occur naturally in real-world situations, significantly lowering the performance compared to that as reported in [1]. Therefore, without solving these problems, which are different from the problems of human recognition, the current deep neural network model is unreliable.

In this study, we analyze the effects of image quality on the performance of a state-of-the-art convolutional network in an image classification task and suggest a novel architecture to overcome the effects. We choose three popular noise types and JPEG lossy compression that can be easily seen in real-world images, which represent low quality. Because there are several factors that generate different types of noise, we use the most common sources of noise that can be seen in the real world—Gaussian, salt-and-pepper (or impulse), and speckle noise. We then investigate the performance degradation in popular neural network models and experiment with the most common input image preprocessing to enhance the performance of the image classification under noisy environments.

We choose the three most common input image denoising methods to make the neural network model robust to quality degradation. The reason for choosing these common denoising methods is that they boost the edges and outlines of an object while suppressing the smaller details. Therefore, we can expect the features from these denoised and outline-enhanced images not to contain any type of quality distortion. We then propose a dual-channel architecture with outline-enhanced input and make a comparison of the small variations in the proposed architecture and its training methods. To generalize our proposed model, we test it with various quality inputs, including an original image dataset, and verify that the model is robust to any type of input quality degradation. We choose three different image denoising methods and verify their effectiveness to strengthen our hypothesis of using an outline-enhanced image.

## II. Related Work

The emergence of the CNN suggests that the neural network model can substitute human labor because the current neural network model has capabilities similar to those of humans. Since its development, there have been a few attempts to explain the vulnerable property of the current neural network model. In an image-understanding task, the convolutional network does not perform as expected when applied to real-world problems. The study by Dodge & Karam [1] showed that image quality, even without visible differences, affected the deep neural network

This work was supported by Basic Science Research Program through the National Research Foundation of Korea (NRF) funded by the Ministry of Education [NRF-2016R1D1A1B03933875].

a negative way and the displayed image output was different from the same image without quality degradation. In the study, the author divided the quality degrading factors into five types: blur, noise, contrast, JPEG compression, and JPEG 2000 compression, none of which disturbed the object identity perceived by a human. What is interesting is that all five types of quality degradation significantly lowered the output labels of the convolutional network. In addition, the study revealed that the filter responses at the upper layers are slightly different compared to the ones from the original undistorted image. Therefore, suppressing the noise or treating the low-quality input image is an unresolved problem that cannot be treated solely by existing convolutional operations.

Proving the effects of the noise in image recognition tasks is a problem not only in CNNs but also in traditional feature extraction methods. The previous paper by da Costa et al. [2] exhaustively researched the effects of different types of noise on the traditional feature extraction methods such as local binary pattern (LBP) and histogram of oriented gradient (HOG). These methods were used extensively in image recognition tasks prior to the use of CNNs. Their study concentrated on the adverse effects of noise on feature extractions of the LBP and HOG methods. They then trained a support vector machine (SVM) classifier for every version of their training set, and each version of a training set had only one type of noise. This setup enabled each model to be specialized for each different type of noise. As a result of this experimental setup and their detailed research, they hypothesized that the performance of a classifier is hindered by a noisy image dataset; the models derived from a clean image dataset are not generalized enough to classify all types of input images and are not robust to noise; and that the study demonstrated image denoising methods. After executing an input image denoising method on a noisy image dataset, the classification performance increased but not sufficiently to obtain the original performance of the original dataset. The authors assumed that this could be due to the loss of details and textures caused by the denoising methods.

Convolutional filters are designed to learn countless numbers of different filters that cannot be designed exclusively by the human hand. A previous study [3] to train filters that can treat various noise on images divided the tasks into two types: near-duplicate detection and image classification. Then the target network was trained with a noisy image dataset. This resulted in the learnt model being resistant to the input noise that the model was trained on. Therefore, this provided evidence that the model could learn the type or pattern of noises and become robust to these. However, the study focused only on pre-defined types of noise to train the model. Moreover, it reduced the accuracy of the original dataset without noise.

There are many known ways to improve the performance of existing models by creating small variations. After capturing an image using real-world sensor, it will acquire certain types of noise. In traditional image processing methods, there are numerous techniques to treat such problems. Moreover, there are trials [4], which designed and implemented image processing methods with denoising and deblurring methods on a deep neural network model. They proposed a combined architecture for image classification tasks, demonstrating a resistant property under sparse light conditions.

In a recent study [5], the author suggested an ensemble model containing specialized models for each quality degradation. This is progress toward a quality resilient model, but still requires every type of expected quality-distorted dataset at the training time. Moreover, they used the VGG16 model [6] which is a shallower network compared to recent network models. More recent deeper neural network models, such as Inception network [7] or Resnet [8], are more appropriate to verify the effectiveness against the original model.

III. EXPERIMENTAL SETUP

*A. Image Preprocessing*

There have been many previous attempts to solve the problem of a noisy or distorted image dataset. For example, the studies [3], [9] focused on the fine-tuning of the neural network model to make it robust to noise, and others, such as [4], attempted to modify the design of the neural network model by adding a novel image processing module ahead of neural network model. We focused on image preprocessing techniques, such as image denoising, to consider real-world images from an imperfect image-capturing device. We will focus on demonstrating that the previous image processing techniques also improve the output of the neural network model. Because the most recent neural network model has very deep layers, over one hundred layers, and has various filters that can extract the various spatial properties of an image, traditional denoising filters may not work on the deep neural network. Therefore, we pick three common denoising methods and attach these before the neural network module.

Nonlocal filtering [10] uses redundant information to suppress noise by performing a weighted average of all pixel values in the image, and by determining the similarities of these pixels to the target pixel. Meanwhile, bilateral filtering [11] performs noise suppressing and smooths images while preserving the edges by replacing the intensity value at each pixel in an image with a weighted average of intensity values from nearby pixels. Lastly, a total variation denoising method [12] reduces the total variation of the abnormal signal to remove any unwanted details and noise while preserving the edges and important details. This is done using the hypothesis that a suspicious signal with excessive detail may have a high total variation. Figure 1 and Figure 2 show the sample output using all three denoising methods. In later sections, nonlocal filtering, bilateral filtering, and total variation denoising are simply mentioned as preprocessing "1", "2", and "3" respectively.

Denoising methods commonly enhance edges and outlines but suppress details and perturbations. Therefore, all three common denoising methods generate an outline-enhanced image dataset. Therefore, the features that are extracted from this dataset have an augmented property of the object shape. If these outline-enhanced features are added to the features from the original image dataset, the robustness of the neural network is improved against quality degradations or small perturbations. Figure 1 illustrates the sample image with the three different methods. Although we can see a clear difference from the original image, all three methods preserved the edges and outline shape. However, the first nonlocal filtering method, tended to wipe out the lines and partial details of the objects inconsistently. These inconsistencies in maintaining the details are reflected in the results and are discussed later.

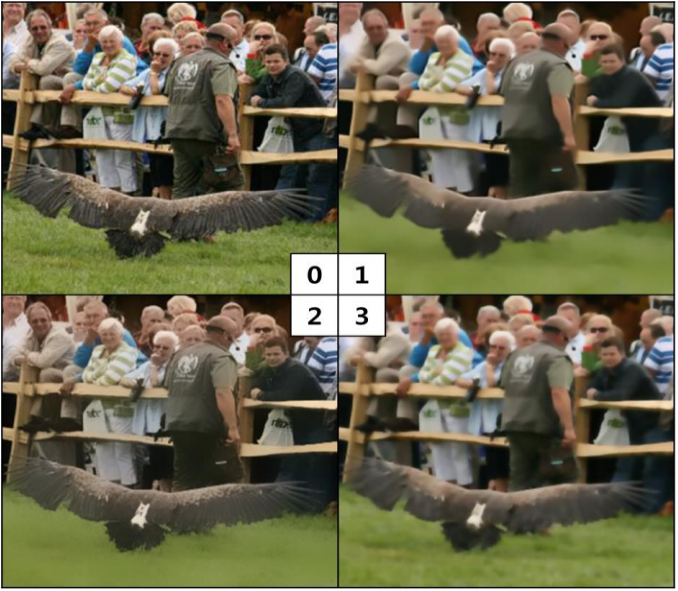

**Figure 1.** Sample images of the four preprocessing methods. The top-left image is the original image. The sample results in the panels were obtained using nonlocal filtering (top-right), bilateral filtering (bottom-left), and the total variation denoising method (bottom-right).

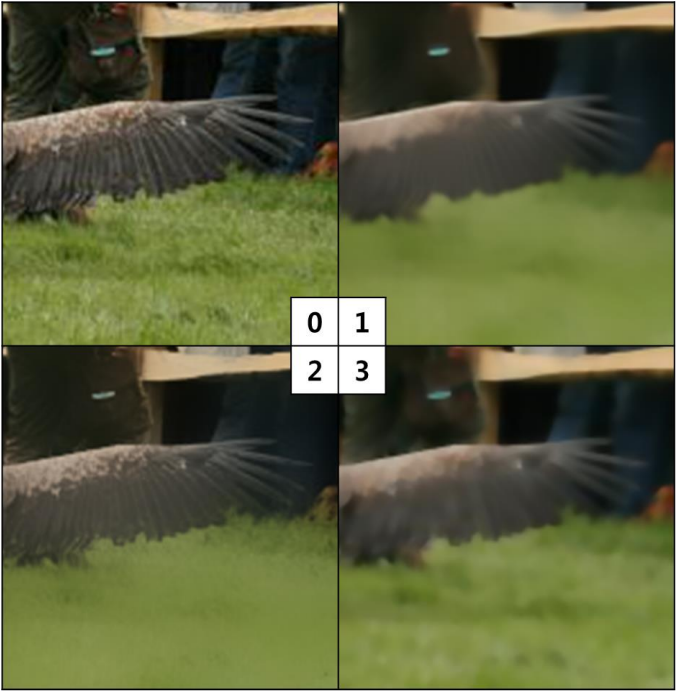

**Figure 2.** Enlargement of the sample image. While all three methods tend to maintain edges and shapes of object, each method has its own property to suppress details and perturbations. If we look into the parts and features of the sample image, compared to the second method, the first method heavily removes the details.

### B. Base Neural Network Model

It is not surprising that models, which have won worldwide competitions in image classification, are generally used in research and many related industries. One of the popular competitions in image classification and object detection is the ImageNet challenge, ILSVRC [13]. In 2014 ILSVRC [13], a novel model [7], having network in network architecture, won the challenge. Shortly afterwards, an improved version Inception-v3 [14] broke records and is the most widely used model today. The Inception-v3 model scores 21.2% top-1 and 5.6% top-5 errors for image classification and attains computational efficiency while expanding and maintaining receptive fields by using asymmetric filters. As shown in Figure 3, the Inception-v3 network consists of micromodules that have many asymmetric filters. The goal of this network in network architecture is to act as a multilevel feature extractor by computing $1 \times 1$, $3 \times 3$, and $5 \times 5$ convolutions within the same module of the network. Because it consists of many different-sized small filters, shown in Figure 4, it is assumed to have a strong representation power leading the model, and should be resistant to small perturbations or quality degradation. Therefore, we use the Inception-v3 model in our experiment. If the preprocessing or denoising methods are effective in this model, it will have capabilities above and beyond the most recent neural networks. We also use the pretrained models, which are publicly available.

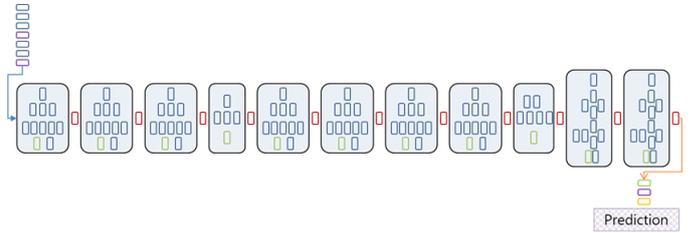

**Figure 3.** Inception-3v network architecture. It is comprised of filters with various shapes including a $1 \times 1$ identity convolutional operation.

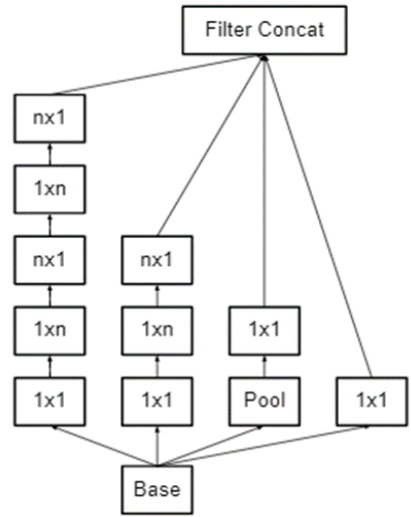

**Figure 4.** Inception inner network modules. This architecture has n size asymmetric filters to aggregate spatial information while preserving representational power. These mini networks replace $n \times n$ convolutions and perform better with less computation than symmetric filters.

## C. Quality Degradation

An imperfect environment during the image acquisition process leads to a ruined image. For example, real-world imaging devices often suffer from many types of noise or blurring. In particular, different environmental conditions produce different types of quality degradation. Under various environmental situations, there can be detrimental factors relating to temperature, light conditions, sensor sensitivity, and improper IOS settings. In addition, noise can appear in the image due to the transmitted signals moving through the electrical circuit. Furthermore, the image quality loss can occur even by compressing the image into a small size JPEG file to obtain additional memory space. We therefore must include these possible quality degradation factors; image noise and JPEG compression are the most common factors. There are several existing models that can reproduce different types of real-world noise.

We used the three most common noise models: Gaussian, speckle, and salt-and-pepper (or impulse) noise. The Gaussian noise model is the most common model that represents general noise in image data. The major causes of Gaussian noise are undesirable environmental factors, such as an unfavorable temperature, poor illumination, or noisy transmission leading to sensor noise. Speckle noise may occur in an active radar, synthetic aperture radar, medical ultrasound, or optical coherence tomography images. Salt-and-pepper noise, also called impulse noise, occurs during digital signal transmission or processing. Note that all these are pixel-wise noise, while JPEG compression generates distorted lines or color fragments as seen in Figure 5. We also used four levels of noise and a JPEG compression rate ranging from a mild to heavy noise level. We chose the levels of the parameters manually. The samples of the four intensities are shown in Figure 6.

## IV. METHOD

### A. Single Model with Image Preprocessing

The current CNN creates less accuracy in the images that contain quality loss or noise, even if it was rarely recognizable. Traditionally image-denoising methods have been used to treat noisy images. Denoising suppresses the small details and perturbations, and enhances the edges. This operation can basically be represented as a blurring of the image, followed by the enhancement of the edges. Therefore, the resultant image emphasizes edges and suppressed details, thereby suppressing the noise in the image.

Although denoising methods have been verified as an effective way to suppress noise, it is unknown if they will be effective when grafted onto a deep neural network. For this reason, we will first perform the input image denoising before grafting it onto the neural network. With this single-model architecture, we empirically prove that training and testing with the denoising method can produce a model robust to some types of noise. Given that it may also be resistant to common quality-distortions, we additionally prepared JPEG compression loss. However, it does not show the general resistant property; it only demonstrates a partial improvement. Consequently, we then propose a dual-channel architecture with a use of outline-enhanced input image.

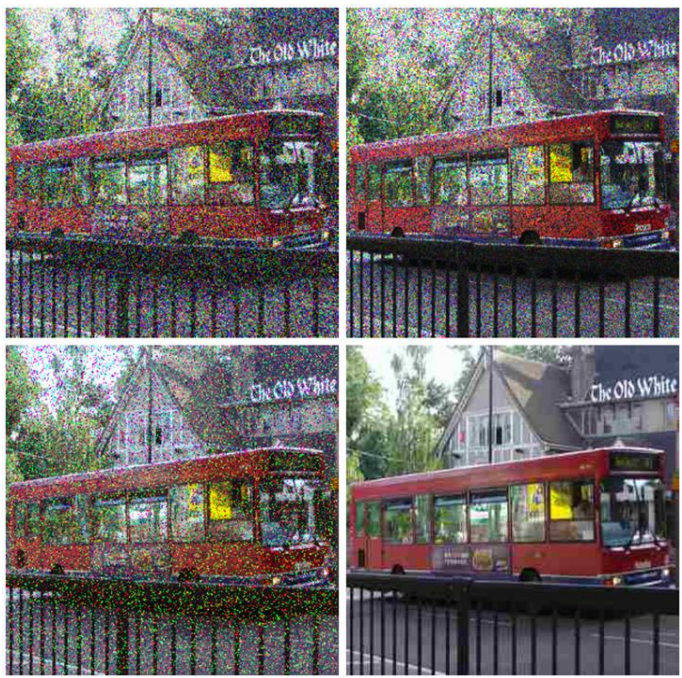

**Figure 5. Four types of quality distortions.** The panels show the samples with Gaussian noise (top-left), speckle noise (top-right), impulse noise (bottom-left), and JPEG compression (bottom-right)

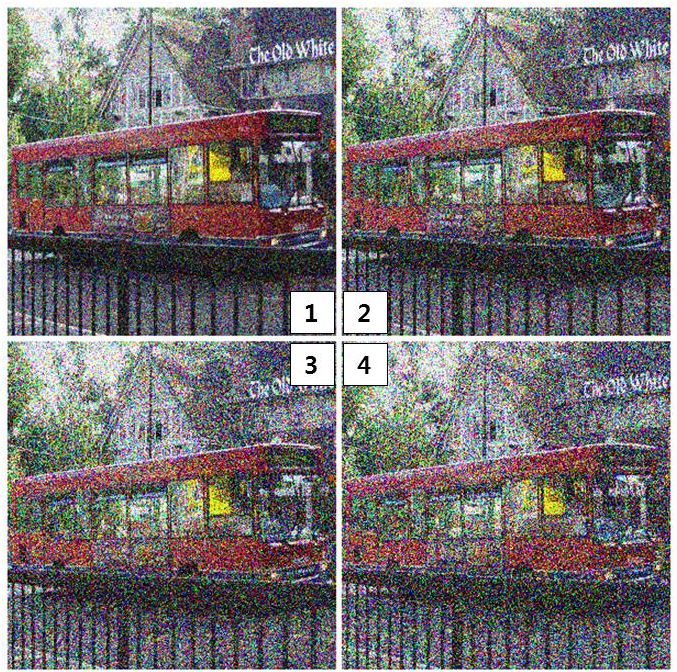

**Figure 6. Four degrees of distortion.** We used four degrees of the distortion level for each of the quality distortions. The sample image is a result of the Gaussian noise with four levels. The numbering refers to the intensity of distortion.

### B. Dual-Channel Architecture

When the target image goes through the CNN module, we can extract various features before going into the fully connected layer. The features include the edges, blobs, or any other details obtained through the convolutional operation. Particularly, in a

2-dimensional (2D) image, convolutional operation is the same as filtering every small region of an image. Therefore, with the deep convolutional network model, we expect the model to have various filters capable of extracting numerous features.

While we can attain many levels of features, from low-level features to high-level features, from an image the obtained features from the outline-enhanced image can be robust to noise and quality degradation. In addition, the denoising method enhances the edges and outlines of an object while suppressing the details or perturbations distributed in the image.

A single model, without a model ensemble, trained and fine-tuned with a noise-suppressed input dataset may reduce the quality of the output from the network on the original image dataset. Although the single model has been proven partially effective for input distortion, we require more stability against all types of quality degradation, including on the original dataset. We designed a new model to maintain the performance on the original image dataset and to utilize the outline-enhanced image that may be robust to all types of noisy input images. This dual-channel model can augment features extracted from the outline-enhanced image, and the architecture is shown in Figure 7. Our ensemble method is similar to the multi-column neural network [15], but differs in that our architecture requires two different inputs from both the outline-enhanced and original image datasets.

The features from the denoised image can be made robust against small perturbations or distortions. To generalize our dual-channel model, we performed broad experiments with all three image-denoising methods that can smooth the image while preserving strong edges as stated in Section III. Then, to enable our model to maintain a high performance on the original image dataset, we also used the original image. Therefore, we pose input image preprocessing on one side and pass the original images on the other side. After the model extracted features from each side of the dual-channel, we merged the two features before they went through the final fully connected layer. Then, we trained the top fully connected layer and fine-tuned the first side, which was adjusted to input image preprocessing. With the dual-channel architecture, the model can provide stable accuracy on any dataset.

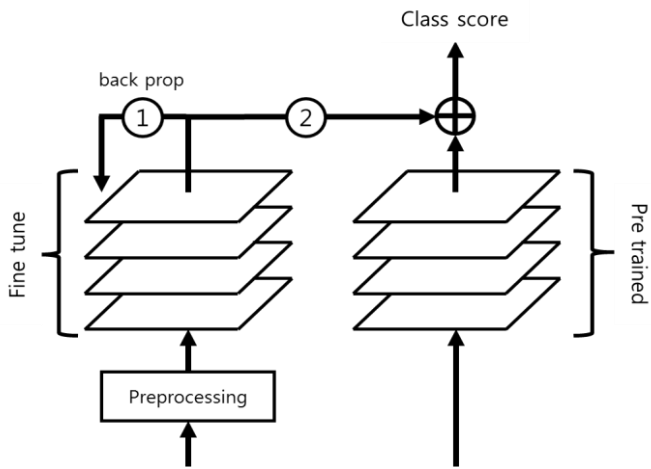

**Figure 7. Dual channel architecture with two inputs. The original model takes the original unprocessed image. The augmented model has a preprocessing module at the bottom of the model, and therefore uses a denoised image as an input.**

There are two different ways of merging the two sides in our architecture. The first method is as follows: before going into the fully connected layer, the dimension of the layer is 2048, and one can concatenate two 2048 2-d vectors to form 4096 dimensional feature maps. The advantage of concatenation is that all the available features are utilized from both sides through the fully connected layer. However, this method creates a computational burden. Therefore, we designed a dual-channel using feature summation. We also compared both methods in this study.

*C. Training with Fine-tuning*

We utilized a pre-trained model for the Inception-v3 architecture, and the input image for our architecture was a denoised image dataset with different properties than those of the original image dataset. Therefore, we needed to fine-tune the entire model. To obtain generality and prove that the preprocessing module makes the CNN model robust to any types of quality distortion, we did not use a noisy image dataset during the training phase. The dual-channel model consists of two CNN models with a concatenation layer and a fully connected layer at the top. Therefore, there were only a few combinations of training the model. The methods were as follows. Firstly, method "1" is training the entire model at the same time. Secondly, method "2" is fine-tuning the left side first, and then training fully connected layer few more times. Lastly, method "3" is training the fully connected layer before fine-tuning the left side model. After testing these methods, we found that the method "3" was the most effective strategy.

For the first one or two epochs, we only trained the fully connected layer for the 50-class classification. Then, to fit the model with the preprocessed input, we fine-tuned the entire model with a low learning rate. The dual-channel model takes two different input images at the bottom. In this case, we only fine-tuned the model that takes an outline-enhanced image as the input, but not the original model.

V. EXPERIMENTAL RESULTS

*A. Dataset*

We performed an experiment using our dual-channel model on the ImageNet 2012 classification dataset to compare with the competition results of ILSVRC [13]. The classification dataset of ImageNet 2012 supports 1000 object classes. We randomly chose 50 out of the 1000 classes for computational efficiency. This is because our model required preprocessing at the training stage and we had to repeat the training epoch many times for the broad experiment. During the training stage, we fed clean images into the original model and preprocessed images into the augmented model. Then, during the testing stage, we imposed four different types of quality distortions previously introduced in section 3 to measure the performance. Note that we did not use the distorted image datasets at the training stage.

*B. Training the Model*

Because our dual-channel architecture consisted of two models, we trained and fine-tuned each of them. Among the few different strategies available to train the two models, we empirically found that method "3" in previous section, training the fully connected layer first and sequentially fine-tuning the first model, is the most effective strategy. While training the entire model made it difficult for the model to converge, it is interesting that training only the fully connected layer was almost as effective as our training strategy. Therefore, training only the final fully connected layer, without fine-tuning the entire model, is also a second effective strategy.

Using the first strategy, we trained the fully connected layer for two epochs on a subset of ImageNet 2012 dataset with an initial learning rate of 0.009. We then exponentially reduced the training rate by dividing by nine at the end of every epoch. Then we fine-tuned the entire model for additional epochs until convergence.

*C. Evaluation Metric*

The top-1 accuracy metric is the most common and widely used metric for image classification. It measures the mean value of the hit ratio out of the total number of test images. The ILSVRC dataset supports 1000 classes of objects. This dataset sometimes includes objects that cannot be defined. Therefore, we used a top-5 accuracy metric as well as the top-1 accuracy metric. We also reduced the total number of classes to 50 by randomly choosing the classes from the 1000 object classes. Therefore, we used the top-1 metric in our experiment.

*D. Results*

Firstly, we tried to verify our hypothesis on a single model fine-tuned with a preprocessed image. Then we performed a number of experiments with quality-distorted image datasets. The quality-distorted image datasets were not used at the training stage. The results of the single model are shown in Figure 8. We measured the model accuracy on the original dataset as well as the quality-degraded datasets. The quality distortions have four levels for the intensity value from one to four. We abbreviate intensity to 'Int.' in Figure 8 and 9 and preprocessing to 'Prep.' in Table 1. Each intensity value is represented as Int. "1", "2", "3", and "4". We show a comparison between the models using the image preprocessing method with the original model, using accuracy ratio graphs with the original model. Therefore, the value for original model is always one.

In Figure 8, the preprocessing methods show better accuracy in several cases. However, the results do not conclusively prove that preprocessing is the better method. Therefore, we designed dual-channel architecture and the results are given in Table 1 and for comparison, the ratio graph is shown in Figure 9. The standard black line is the original model without any preprocessing.

The results of the single-model experiment suggest that it is not an appropriate model to use for image preprocessing of a noisy image dataset. Moreover, the single model setup reduces the performance using the original image dataset causing the model to be underfit. However, the dual-channel model, which was expected to have additional features from the outline-enhanced image, shows better performance than the original model on the noisy image dataset. Moreover, the dual channel model shows a comparable performance on a clean image dataset. The results of this experiment show that the model is robust to any types of input image distortions, even if the distorted image is not used at the training stage.

In Figure 9, the preprocessing method "1" produced less improvement because it removed important details of the object inconsistently and, unlike the other two methods, diluted the key edges of an object. The preprocessing method "2" with bilateral filtering showed the best results as with the sample image.

As stated earlier, there are two different ways of merging the

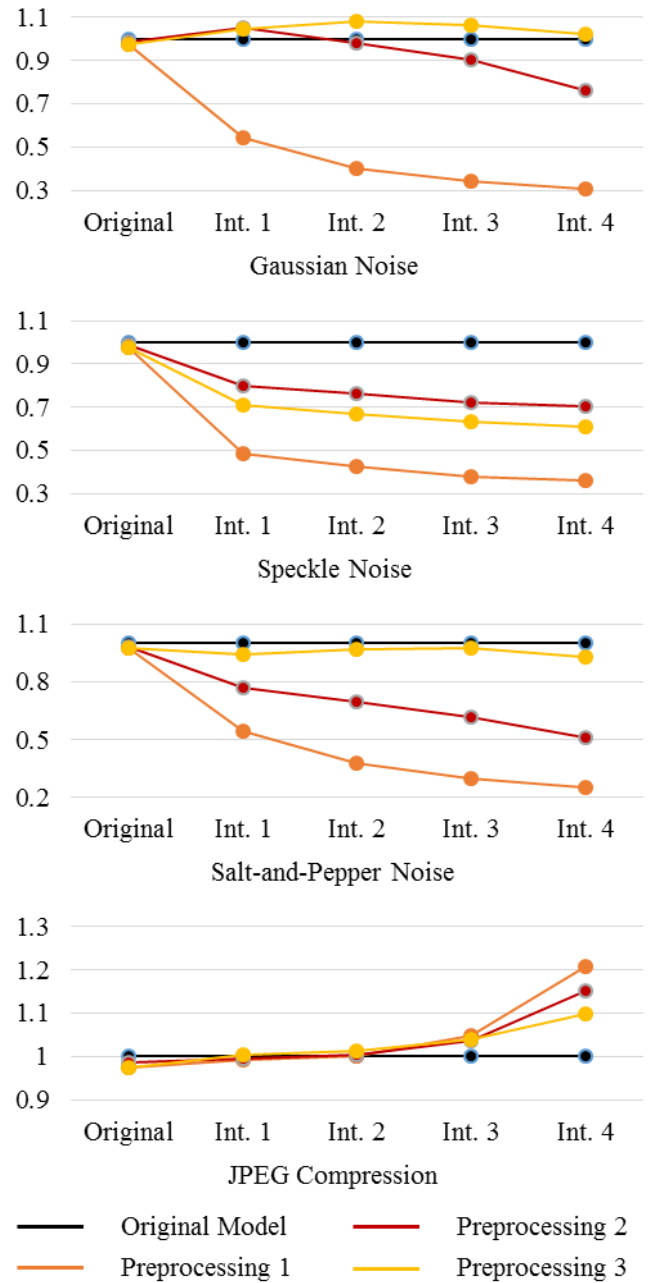

**Figure 8. The result of a single model. The results are shown by ratio to the accuracy of original model. Single model architecture is only partially effective against quality distortions. Preprocessing 1 is nonlocal filtering, 2 is bilateral filtering, and 3 is total variation denoising method**

Table 1. The results of the dual-channel architecture. We used three models with three image-preprocessing methods against four distortions. Table (a) has Gaussian noise; (b) has Speckle noise; (c) has Salt-and-pepper noise; and (d) includes JPEG compression.

| Intensity | 1 | 2 | 3 | 4 |
|---|---|---|---|---|
| Original | 64.74 | 48.15 | 35.53 | 27.59 |
| Prep.1 | 64.98 | 48.48 | 36.82 | 27.93 |
| Prep.2 | 68.79 | 53.57 | 40.84 | 30.7 |
| Prep.3 | 68.44 | 51.37 | 38.87 | 27.7 |

(a)

| Intensity | 1 | 2 | 3 | 4 |
|---|---|---|---|---|
| Original | 66.70 | 57.53 | 51.36 | 47.19 |
| Prep.1 | 66.71 | 58.42 | 53.13 | 48.21 |
| Prep.2 | 69.11 | 60.55 | 55.38 | 49.85 |
| Prep.3 | 66.59 | 58.38 | 51.89 | 47.69 |

(b)

| Intensity | 1 | 2 | 3 | 4 |
|---|---|---|---|---|
| Original | 70.95 | 54.88 | 40.54 | 30.36 |
| Prep.1 | 71.56 | 55.15 | 42.03 | 32.1 |
| Prep.2 | 72.88 | 58.78 | 46.16 | 33.62 |
| Prep.3 | 72.92 | 56.90 | 43.15 | 33.74 |

(c)

| Intensity | 1 | 2 | 3 | 4 |
|---|---|---|---|---|
| Original | 88.22 | 87.90 | 82.61 | 63.90 |
| Prep.1 | 88.46 | 87.94 | 83.38 | 66.15 |
| Prep.2 | 89.06 | 88.46 | 84.06 | 67.87 |
| Prep.3 | 88.78 | 87.78 | 84.34 | 66.19 |

(d)

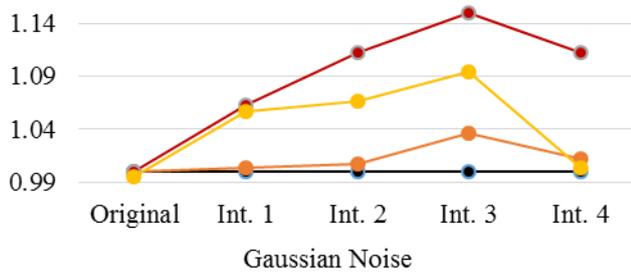
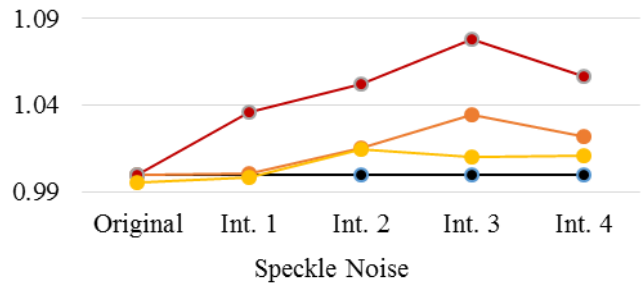
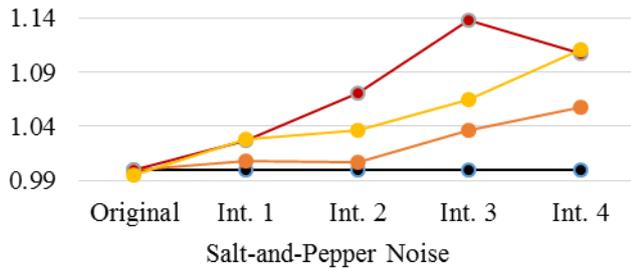
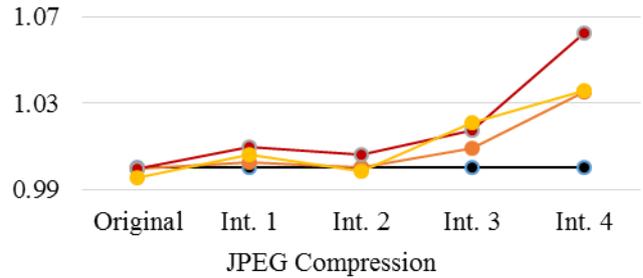

Figure 9. The result of dual-channel architecture compared to the original model. The ratio graph visualize the performance gain of our architecture. In general, our dual-channel models with an image preprocessing module records better accuracy compared to the original model.

two sides in our architecture. One is the summation of two features from the dual-channel; the other is concatenation of the two features. Table 2 summarizes a comparison of these two.

Table 2. Comparison of two different merging methods. The intensity level is set to "2" and preprocessing bilateral filtering is applied.

| Distortion | Gaussian | Speckle | Impulse | JPEG |
|---|---|---|---|---|
| Concatenation | 53.57 | 60.55 | 58.78 | 88.46 |
| Summation | 47.85 | 52.81 | 56.82 | 88.34 |

The results show that the feature concatenation is far more effective in our architecture than feature summation.

## VI. DISCUSSION AND CONCLUSION

In this study, we introduced two different solutions to enhance the performance of image classification. In a real-world classification task, a deep neural network often indicates reduced accuracy. The problem is that real-world images often include noise or quality loss. We proposed an input image denoising method and dual-channel architecture to achieve a stable performance of image classification in a real-world task. We chose the most common image denoising methods to investigate the integrity of the input image preprocessing. However, a preprocessing method on its own did not produce any significant improvement. The preprocessing method proved successful only in the form of our dual-channel structure.

The dual-channel model contains extra features from the outline-enhanced image that are expected to be robust to quality distortion. Our hypothesis was based on the intuition that outlines, edges, and color components of the object are enhanced even after quality loss. Moreover, because our architecture utilized an outline-enhanced image as an augmented feature as well as the original image, our dual-channel architecture did not reduce the accuracy of the original image dataset while achieving better records of all types of quality distortion. This is meaningful in that the model did not depend on the type of noise used in the training phase. Therefore, we interpret our architecture as a generalized model for any types of quality loss.

Our current dual-channel architecture for image classification is comprised of a neural network model and a separate preprocessing module. Therefore, in future, we intend to design architecture with end-to-end learning structures and define other preprocessing modules as well. In the current research, we used well-known denoising methods as a preprocessing module to emphasize the outlines and edges and as a method to reinforce the specific features. If we are successful in developing a customized preprocessing module, better results are expected. In this study, the target task was the image classification within the image. In future, we aim to apply our architecture to semantic segmentation [16] or object detection [17] for further generalization.